# Clustering by Deep Nearest Neighbor Descent (D-NND): A Density-based Parameter-Insensitive Clustering Method

Teng Qiu
(qiutengcool@163.com)

Yongjie Li
(liyj@uestc.edu.cn)
University of Electronic Science and
Technology of China, Chengdu, China

**Abstract:** Most density-based clustering methods largely rely on how well the underlying density is estimated. However, density estimation itself is also a challenging problem, especially the determination of the kernel bandwidth. A large bandwidth could lead to the over-smoothed density estimation in which the number of density peaks could be less than the true clusters, while a small bandwidth could lead to the under-smoothed density estimation in which spurious density peaks, or called the "ripple noise", would be generated in the estimated density. In this paper, we propose a density-based hierarchical clustering method, called the Deep Nearest Neighbor Descent (*D-NND*), which could learn the underlying density structure layer by layer and capture the cluster structure at the same time. The over-smoothed density estimation could be largely avoided and the negative effect of the under-estimated cases could be also largely reduced. Overall, D-NND presents not only the strong capability of discovering the underlying cluster structure but also the remarkable reliability due to its insensitivity to parameters.

## 1  Introduction

Cluster analysis is a popular and powerful data analysis tool in diverse fields as science, engineering and business (1-3). Based on the pair-wise similarities, It is used to divide into groups the massive dataset, e.g., the gene expression profiles, the images and texts on the internet, or users' consuming data. By clustering, the raw data could be effectively organized and the underlying group relationships among data points could be discovered. The raw data could thus become useful and informative to us.

Among various ways of clustering, the idea of density-based clustering never loses its appeal (4-10). Like K-means (11), the most popular clustering method, most of the density-based clustering algorithms are *(i)* simple and effective, *(ii)* with relatively low time complexity and thus eligible for processing large datasets (3). Besides, unlike K-means, most of density-based clustering algorithms are *(iii)* able to detect arbitrarily shaped clusters, *(iv)* robust to noise or outliers, *(v)* insensitive to the initialization and *(vi)* of no need to specify the cluster number in advance. Besides, density-based clustering is *(vii)* quite intuitive, since, for instance, the input data points $\Psi = \{x^i\}_{i=1}^{N}$ (Fig. 1A) are usually associated with a mountain-like density function with multiple density peaks as Fig. 1 B shows. Such density function reflects the distribu-tion of the data points. Obviously, the points associated with the same mountains could be grouped into same clusters and the points corresponding to the density peaks represent the cluster centers.

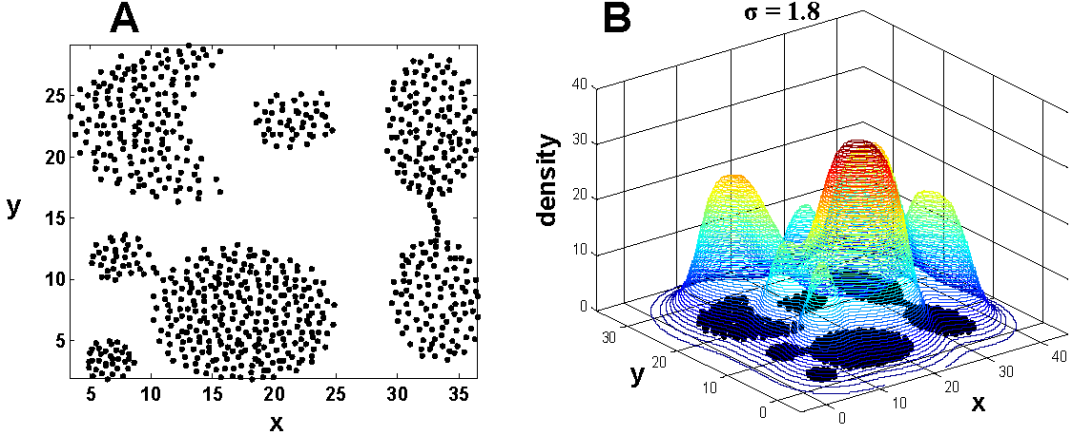

**Fig. 1. An illustration for the well estimated density function.** Left: input dataset. Right: the estimated density function.

Therefore, in order to divide the whole data set into clusters, two principal Questions need to be answered:

- **Q1**: *how to estimate the underlying density function?*
- **Q2**: *based on the estimated density function, how to do the clustering assignment? or how to separate different "mountains" apart?*

These two questions, however, are not as easy to be simultaneously solved as they appear. In the following, we will introduce several density-based clustering methods which are, directly or indirectly, consciously or unconsciously, involved in solving the above Questions. Note that the methods are not introduced in a Chronological Order.

## 1.1 DENCLUE

DENCLUE (6) should be a representative and intuitive density-based clustering method. For the Question Q1, the underlying density function is simply approximated by the sum of certain kernel functions centered at each point,

$$f(x) = \sum_{i=1}^{N} K_\sigma(x - x_i). \tag{1}$$

For square wave kernel, $K_\sigma(x) = 1$ if the distance $\|x\| \leq \sigma$; otherwise, $K_\sigma(x) = 0$; For Gaussian kernel, $K_\sigma(x) = \exp(-\|x\|^2/2\sigma^2)$. $\sigma$ is called the bandwidth, a scalar parameter usually set by users. Note that for simplicity, the density images shown in this paper are all computed based on the kernel: $K_\sigma(x) = \exp(-\|x\|/\sigma)$.

Although DENCLUE provides an easy and straightforward solution for the ques-tion Q1, whereas making it hard for the question Q2. This is just opposite to the model-based clustering methods typically as GMM (12) and t-distribution Mixture Model (13, 14). For instance, GMM assumes that the data points are sampled from a probabilistic density function, a combination of $K$ ($\ll N$) multivariate Gaussian functions. Once solving the parameters in GMM, the clustering assignment (i.e., the question Q2) can be easily solved by the Bayesian criterion. Nevertheless, it is well-known that it is hard to solve GMM analytically and the iterative solution EM (15) is usually used as an alternative to solve the parameters, whereas GMM still involves two *open* problems, i.e., sensitive to initialization and hard to specify component $K$ in advance.

For DENCLUE, obviously, the clustering assignment cannot be simply solved[1] by the Bayesian criterion like GMM. Instead, DENCLUE uses the classical **Gradient Ascent (*GA*)**, also known as Steepest Ascent or the "hill-climbing" method, to identify the density peaks. Specifically, each point $x$ "shifts" in the direction of gradient ascent by the following iteration:

$$x^{t+1} = x^t + \lambda \cdot \nabla f(x^t) / \| \nabla f(x^t) \|, \tag{2}$$

where $\lambda$ is the shifting step, $x^t$ and $x^{t+1}$ are the current and next locations, respectively. The iteration was designed to stop at $t \in N$ if $f(x^{t+1}) < f(x^t)$. Then $x^t$ was taken as the so-called density-attractor (approximately the density peak), and each density-attractor together with the attracted points were assigned into the same cluster. Besides, a threshold was also suggested to judge whether the attractor is significant, which is useful for distinguishing the outliers, since they usually have low densities. Overall, DENCLUE provides a simple, intuitive and effective way to solve the question Q2 by using GA, whereas, in our opinion, GA also brings at least the following nontrivial problems:

- ✧ *P1*: *fixed step $\lambda$*. This is an inherent problem for GA. That is, a small step length could make the iteration slow to converge, while a large $\lambda$ could make the iteration oscillate around certain density peak.
- ✧ *P2*: *iterative process*. This is theoretically time-consuming, despite certain acceleration strategy available.
- ✧ *P3*: *limited applications*. GA can be applied only to the numerical vector data, while there are many datasets in practice that contain categorical vectors or directly the pair-wise dissimilarity rather than vectors.
- ✧ *P4*: *sensitive to the kernel bandwidth $\sigma$*. Compared with Fig. 1B, a small $\sigma$ could make the estimated density function under-smoothed (Fig. 2A), with spurious density peaks (or called the "ripple noise"), which would lead to the "over-partitioning" clustering results. For instance, both points a and b in Fig. 2A would become two different density-attractors or cluster centers for the same cluster they both belong to, and in fact, in Fig. 2A, there exist at least three clusters that have the same risk to be over-partitioned. In contrast, a large $\sigma$ would make the estimated density over-smoothed (Fig. 2B), which will lead to the "under-partitioning" clustering result.

---

[1] Since, unlike GMM, the number of components here is obviously the number of data points, far more than the cluster number.

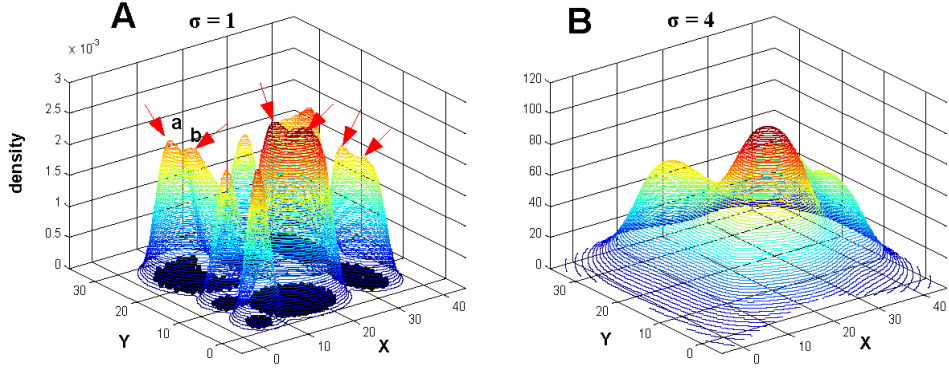

**Fig. 2. Illustrations for the bad estimated density functions.** Left: under-smoothed; right: over-smoothed.

Nevertheless, DENCLUE also has some good properties (6). For instance, it is insensitive to noise. This is not hard to understand, since the noise can hardly make significant effect on the overall density function. Besides, DENCLUE has a close relationship with both DBSCAN (5), another popular density-based clustering method, and K-means, the most popular partition-based clustering method. That is, DBSCAN turns out to be the special case of DENCLUE when DENCLUE uses the uniform spherical kernel (also called the square wave kernel); the globally optimal clustering results of K-means are the same with the clustering results of DENCLUE, provided that the cluster numbers are identical.

### 1.2 Mean-Shift

Researchers further found that in this particular problem the following iteration can also locate the modes (density peaks) as what GA does:

$$x^{t+1} = \frac{\sum_{i=1}^{N} x_i K_\sigma(x^t - x_i)}{\sum_{i=1}^{N} K_\sigma(x^t - x_i)}. \qquad (3)$$

We can see from this iteration expression that (i) the next location $x^{t+1}$ is simply computed by the weighted mean around the current location $x^t$, and that (ii) the fixed step parameter $\lambda$ in GA is avoided here. In fact, the step length here (from $x^t$ to $x^{t+1}$), denoted as $\tau$, is adaptive to the local distribution. The denser the area is, the smaller $\tau$ would be. This method is the so-called *Mean-Shift*, first proposed in 1975 by Fukunaga and Hostetler (16), developed and brought into notice again in 1995 by Cheng (17), and become popular since 2002 by Comaniciu and Meer (18). Mean-shift proved (17, 18) to be equivalent to the GA on the density function $\tilde{f}(x)$ defined as follows,

$$\tilde{f}(x) = \sum_{i=1}^{N} \tilde{K}_\sigma(x - x_i), \qquad (4)$$

where $\tilde{K}_\sigma(x)$ is the so-called shadow of $K_\sigma(x)$, and for normal kernel, they are of the same expression. Mean-Shift is guaranteed to converge by using certain kernels (e.g., normal and Epanechnikov kernels) (18). However, Mean-shift is still of iterative nature in searching the modes and thus still theoretically time-consuming. It is for this reason that many efforts (18-23) have been made to accelerate it. Besides, it is also easy to find that, like DENCLUE, Mean-shift is only suitable for numerical data and sensitive to the kernel parameter $\sigma$, despite the fact that the density function is not needed to be computed.

### 1.3 Graph-based Gradient Ascent (*Graph-GA*)

Besides Mean-Shift, there is in fact another variant of GA. Unlike Mean-Shift, this method, proposed in 1976, is implemented based on the graph theory instead of numerical analysis, and thus, we call this method the Graph-based Gradient Ascent (***Graph-GA***) (4). Graph-GA proved (4) also equivalent to the GA in terms of the kernel function $f(x)$. Graph-GA approximates the gradient at each point *i* with referring to its neighbors as follows,

$$\hat{\nabla} f(x_i) = \max_{j \in \eta_\theta^i} \frac{\rho_{ji}}{d_{ji}}, \qquad (5)$$

where $\rho_{ji} = \rho_j - \rho_i$ denotes the density difference, $d_{ji}$ the distance between points *i* and *j*, and $\eta_\theta^i$ the neighbors of point *i* within the radius $\theta$. However, the value of the estimated gradient was not used. Instead, based on the above expression for gradient, what the researchers care most was the neighbor that corresponds to the maximum output. This particular neighbor, denoted as $I_i$, also called the parent node of point *i*, was thus defined as

$$I_i = \arg\max_{j \in \eta_\theta^i} \frac{\rho_{ji}}{d_{ji}}. \qquad (6)$$

If we treat each point as a node in the graph and link each node to its parent node by a directed edge, then this would result in a directed and usually unconnected graph, with several separated sub-graphs. Each sub-graph is a directed tree[2] containing a root node. All the root nodes approximately correspond to the density peaks. Based on this graph (functioning like a map), the next process of identifying the attractor (i.e., the root nodes here) of each node could simply follow the "directions" of the edges on the graph. In other words, the next computation only involves a serial of identification of the parent node $I_i$, the parent node of node $I_i$, etc., the computation time of which is almost negligible, as compared with GA or Mean-shift in which each step of approaching the mode involves the same computation as the first iteration. For this reason, Graph-GA can be more efficient than GA and Mean-Shift. Besides, Graph-GA is not constrained by the attributes of the data and can directly take the dissimilarity as the input. However, like Mean-shift, Graph-GA is also sensitive to the kernel-like pa-rameter *θ*. This is not hard to understand, since Graph-GA is just a graph-based implementation of GA and thus they are in essence the same. Besides, Graph-GA itself involves nontrivial process of avoiding cycle in the Graph, and is somewhat sensitive to the order in which the data points are processed, which, in our perspective, should be largely due to the way utilized to avoid cycle.

---

[2] While it is hard to say whether it can be guaranteed to be an in-tree. This is left into question.

*Brief summary:*

*DENCLUE uses the original form of **GA**, which brings in four non-trivial problems (P1~P4); **Mean-shift** changes the way of iteration and consequently avoids the fixed step problem of GA, whereas still suffering from P2~P4; **GRAPH-GA** solves the first three problems while the last one still remains.*

*In the following, we will introduce the new progresses on this issue. First, we will introduce a novel method proposed one year ago, which we call here the **Decision Graph** (**DG**). DG is an extremely simple and effective method, not only successfully solving the problems P1~P3, but also largely reducing the negative effect of the "ripple noise"(the under-smoothed case) in the fourth problem. Besides, as a whole, we will also introduce a novel method previously proposed by us, very similar to DG, called the **Nearest Descent** (**ND**), together with other new progresses we have made from ND to the **Nearest Neighbor Descent (NND)**, and the **Hierarchical Nearest Neighbor Descent (H-NND)**. Both ND and H-NND have comparable advantage with DG, while NND, like Graph-GA, can only solve the first three problems, but it can serve as a tie for the old methods and new ones and a pivotal element in the method proposed in this paper.*

## 1.4  Decision Graph (DG)

In 2014, Rodriguez and Laio (7) proposed a simple and effective method which can fast determine the density peaks in a 2D scatter plot, called the Decision Graph, in which the density peaks pop out and can be interactively determined by users. This Decision Graph is plotted based on two effective features or variables they discovered. One is the local density $\rho_i$, and the other is a special distance $\delta_i$, defined as the minimum distance of point *i* to other points in the dataset of larger density:

$$\delta_i = \min_{j \in \Psi} d_{ji} \cdot \varphi(\rho_{ji}) \quad \text{for all} \quad i \in \Psi, \tag{7}$$

where $\varphi(x) = 1$, if $x > 0$; otherwise, $\varphi(x) = +\infty$. They found that only the density peaks have large values in both features, $\rho_i$ and $\delta_i$, and thus would pop out in the 2D feature space coordinated by these two variables (Fig. 3 A or C) and could be interactively identified (the red box in Fig. 3A or Fig. 3C records the user's operation of identifying those popping out points). Since the density $\rho_i$ for each point in Fig. 3C is just the case in Fig. 2A (i.e., the case with ripple noise), the corresponding clustering result in Fig. 3D clearly reveals that DG largely reduce problem brought by the "ripples". In other words, ***only the valid density peaks are selectively shown in the Deci-***

*sion Graph*. For instance, for points a and b in Fig. 2A, only point a will pop out in the decision Graph and be taken as cluster center. The reason is that, despite the high density $\rho$ for both points a and b, the distance variable $\delta$ for point b is small (just the distance between a and b) while the distance variable for point a is much larger (the distance between points a and the nearest point of higher density in other cluster). Note that in Fig. 3C we only select 7 most salient points, while it is obvious that there is another point outside the red box also saliently popping out. If we also enclose that point into the red box, this would make one cluster (the cluster in green in Fig. 3D) over-partitioned. Nevertheless, the negative effect of the "ripple noise" would still be largely reduced, since the number of the clusters that could be over-partitioned could be reduced from 3 (at least) to 1.

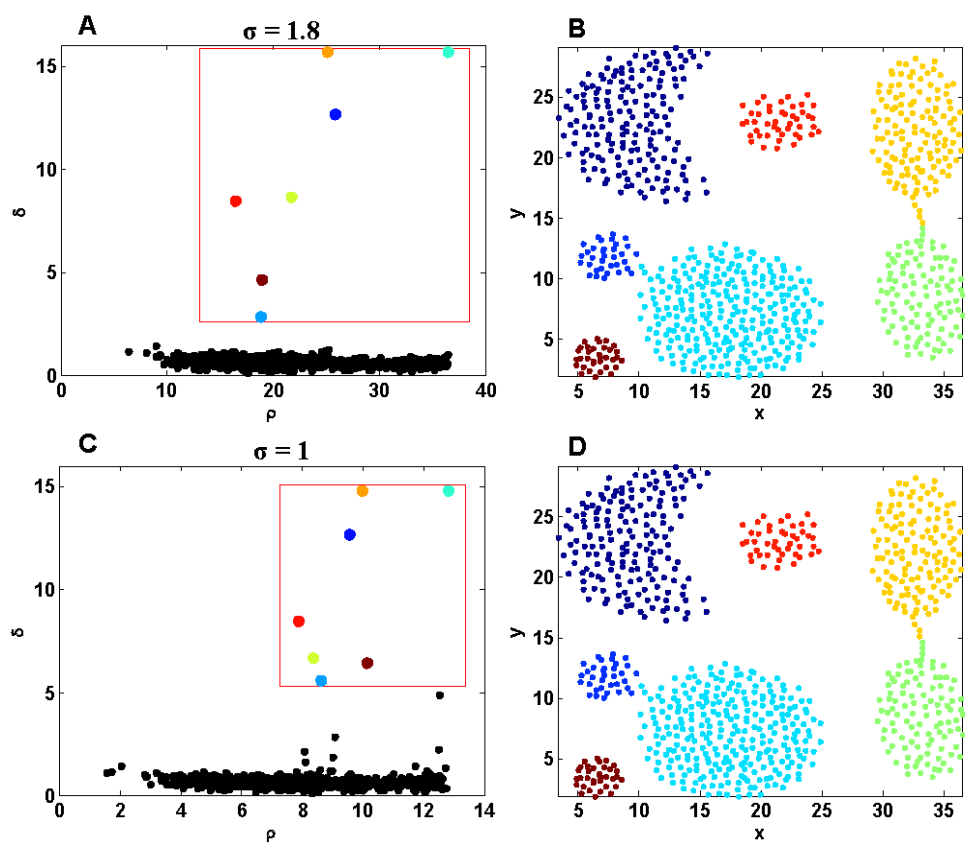

**Fig. 3. Illustrations for DG.** Left column: Decision Graphs obtained in different parameters. Right column: the corresponding clustering results.

Overall, the success of DG is not only because it could provide a fast, non-iterative and interactive way to identify the density peaks, since both Graph-GA and the following method NND could do it as well. But also because it could resist the disturbance of the invalid density peaks in the "ripples" (the under-smoothed case), which was also discussed by Rodriguez and Laio in the end of their paper. The latter property contributes to the fact that the sensitivity of DG to kernel bandwidth is largely reduced.

One thing worth noting is that the interactive operation in DG could in general enhance the reliability of the whole process, since it is believed (24) that the participation of users in the visualized environment could contribute to the reliable exploration of the unknown data's world. However, when the bandwidth is too large[3] (corresponding to the over-smoothed density), behind the interaction operation (Fig. 4A), there would still exist the risk of generating misleading results (Fig. 4B, the clusters in blue are under-partitioned). Besides, such risk would also exist when there is no clearly separated cluster structure in the dataset, since in this case it would be hard to obtain an explicit Decision Graph with some saliently popping-out points.

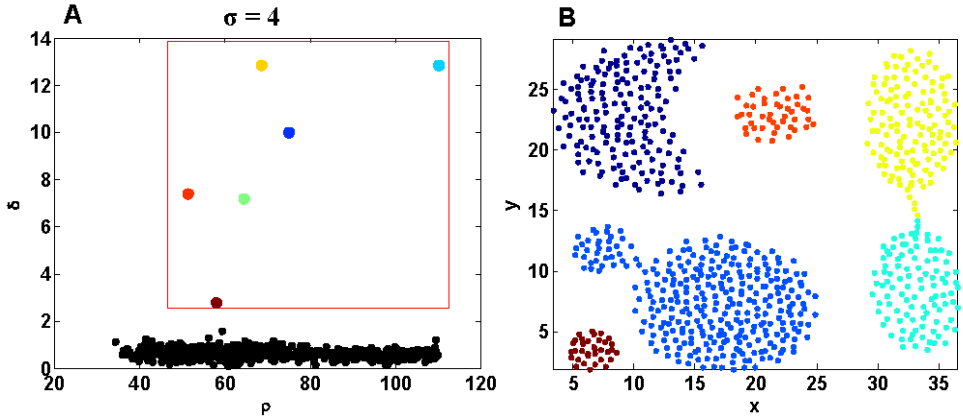

**Fig. 4. A false case of DG.** Although the Decision Graph (left) has very salient pop-out points, the corresponding clustering result (right) is under-partitioned (see the cluster in blue).

## 1.5 Nearest Descent (ND)

Also in 2014, we opened our physically inspired method (8), while, instead of using the concept of gravitational force between particles to explain their movement which would make the evolution for a *particle system* extremely complex[4], this method was inspired by another idea[5] which reckons that particles curve the space, and in turn in the curved space particles tend to move (i.e., Einstein's view of the universe). Specifically, the curved space can be approximately viewed as the potential profile in Fig. 5B and all points (treated as particles) directly lie on it. The vertical axis in Fig. 5B is called the potential $P$. Actually, in terms of computation, the potential is just negative to density, that is,

$$P(x) = -f(x) = -\sum_{i=1}^{N} K_\sigma(x - x_i),  \qquad (8)$$

and the potential profile in Fig. 5B is inverse to the density profile in Fig. 2A (the ripple noise case). Since this potential axis is not of the same physical attribute with the other two coordinate axis, we can view this curved space in Fig. 5B as a 2.5D space.

---

[3] Although DG is insensitive to the parameter, the range of the values that can lead to a good performance is relatively narrow.
[4] In fact, this complexity is to some degree revealed by the affinity propagation (AP) clustering, although AP is not based on the gravitational force (instead, the "massage" passing process for the pair-wise points).
[5] http://einstein.stanford.edu/SPACETIME/spacetime2.

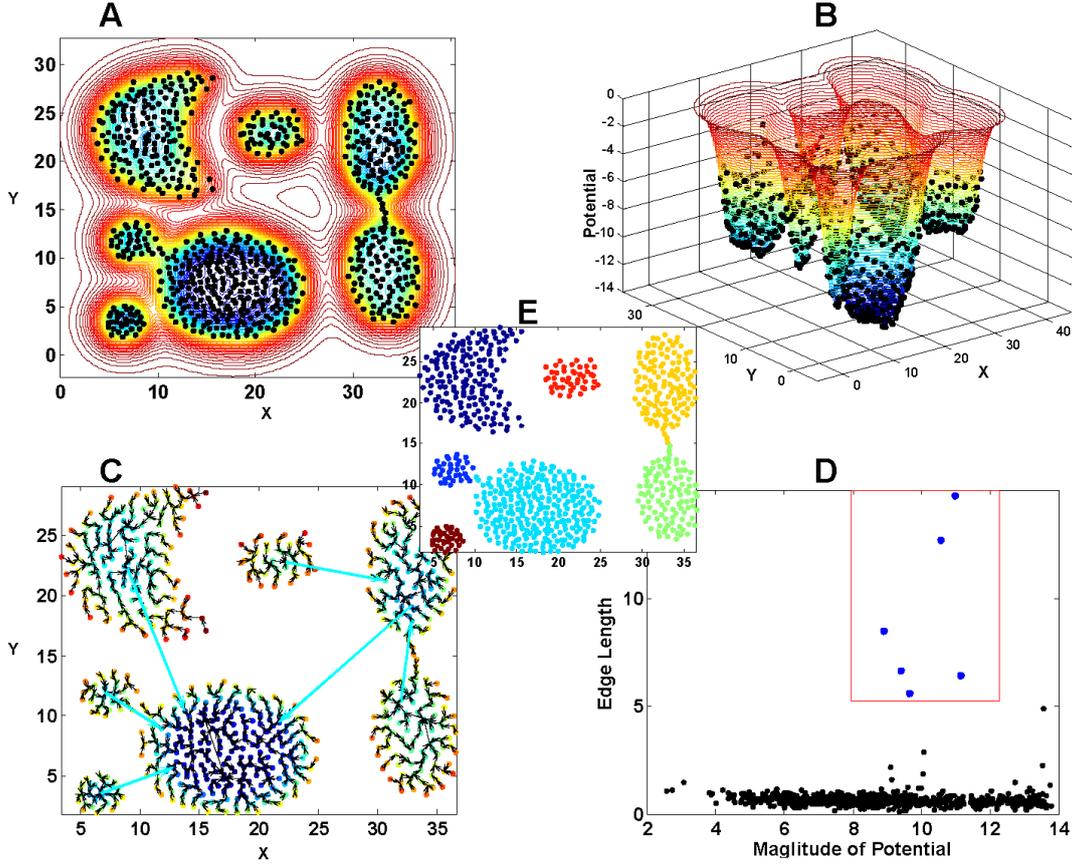

**Fig. 5. An illustration for ND. (A)** The estimated potentials on a 2D dataset. The bluer the areas appear, the lower the potentials are. **(B)** An equivalent representation to (A) in a 2.5D space where the potential is represented by an additional axis. **(C)** The IT data structure constructed by ND. The colors on nodes denote the potentials. The edges in cyan denote the redundant edges that need to be removed further. **(D)** The Decision Graph used to determine the redundant edges. The identified pop-out points (in blue) correspond to the start nodes of those redundant edges. **(E)** The clustering result after removing the redundant edges.

Also, we can use the original 2D physical space to approximately represent the curved space, as Fig. 5A shows, in which the different "curvature" in space can be approximately revealed by the uneven potentials in space (different colors denote different potentials there). Consequently, due to this uneven curvature in space (e.g., Fig. 5B) or the uneven potential distribution in space (e.g., Fig. 5A), we can almost "feel" the evolution tendency of the particle system, that is, each particle tends to "descend" in the descending direction of potential. This tendency was simplified as an algorithm named the **Nearest Descent** (***ND***), where "Descent" refers that each particle "descends" in the descending direction of potential. Specially, *ND means that each point "descends" to the nearest one (denoted as $I_i$) in the descending direction of potential*. And accordingly, $I_i$, also called the parent node of point $i$, is defined as

$$I_i = \arg\min_{j \in \Psi} d_{ji} \cdot \varphi(-P_{ji}) \quad \text{for} \quad i \in \Psi - r, \tag{9}$$

where $P_{ji} = P_j - P_i$ ($P_{ji} = -\rho_{ji}$), $\varphi(x)$ is defined in the previous equation (i.e., $\varphi(x) = 1$, if $x > 0$; otherwise, $\varphi(x) = +\infty$ ), and node $r$ is the one[6] with the globally lowest potential in the dataset. Like Graph-GA, from a global perspective, *one time of independent "hopping" of each point to its parent node will lead to a graph structure*, as Fig. 5C shows. However, unlike Graph-GA, ND is guaranteed to make all data points organized into a fully connected graph, with a special name called the **in-tree** (***IT***) in Graph theory (25). IT, also called in-arborescence or in-branching, is a directed graph that meets: (i) only one node (also called the root node) with outdegree 0 (i.e., no directed edge started from it); (ii) any other node with outdegree 1; (iii) no cycle; (iv) fully connected. We have proven that such IT structure is guaranteed to be generated by ND.

Actually, this intermediate result IT, as in Fig. 5C, shows something surprising and exciting that we have not expected at the beginning. For instance, in this IT structure, we can see two significant features. (i) all clusters have already been captured in it except a small number of redundant edges. For this reason, this clustering problem is then reduced to the classical edge-removing problem, alike the Minimal-Spanning-Tree (MST) based clustering (26). (ii) it is easy to find that the redundant edges are very distinguishable from other edges and thus would be very easy to be removed, which is, however, quite unlike MST-based clustering. For MST, among the close clusters or the clusters contaminated by noise, there would exist short-linked redundant edges that are hard to be determined. In contrast, for IT (Fig. 5C), we can determine the redundant edges, for instance, simply by ranking the edge lengths in decreasing order and choose the six longest ones. The clustering result in Fig. 5E shows that ND could also largely reduce the negative effect of the "ripple noise".

In fact, as Fig. 5D shows, DG can also serve as an effective method to determine the redundant edges in IT (we denote this method as DG-cut here). This is because, according to the 2nd requirement of IT, each node $i$ (except the root node) has one and only one directed edge started from it, i.e., edge $(i, I_i)$, we can thus use the length of this edge $(i, I_i)$, together with the magnitude of potential on node $i$, as two variables of each node $i$, so as to obtain a similar Decision Graph like Fig. 3A. In fact, from the technical point of view, ND and Rodriguez and Laio's DG happen to be, in essence, the same, despite their different backgrounds (ND is physically inspired) and implementations (ND is graph-based).

> *In fact, we can view Rodriguez and Laio's DG as the counterpart of ND, i.e., the **Nearest Ascent (NA)**, where "ascent" refers to the ascending direction of density. In other words, NA means that each point "ascends" to the nearest point in the ascending direction of density.*

---

[6] The special case with more than one such points has been considered in ND.

## 1.6 Nearest Neighbor Descent (NND)

Later on, we proposed a method, called the Nearest Neighbor Descent (*NND*) (9). Unlike ND, NND requires that each point descends to its nearest ***neighbor*** in the descending direction of potential. In other words, the parent node $I_i$ of node $i$ in NND is constrained by the neighborhood relationship, as follows,

$$I_i = \arg\min_{j \in \eta_\theta^i} d_{ji} \cdot \varphi(-P_{ji}) \quad \text{for} \quad i \in \Psi - Y, \tag{10}$$

or,

$$I_i = \arg\min_{j \in J_i} d_{ji} \quad \text{for} \quad i \in \Psi - Y, \tag{11}$$

where $Y = \{i \mid J_i = \varnothing\}$ and $J_i = \{j \mid P_{ji} < 0, j \in \eta_\theta^i\}$. $Y$ denotes the local minimum points (or the root nodes in the generated graph). $J_i$ is called the ***candidate*** parent nodes of node $i$. By the neighborhood constraint of the parent node, NND can prevent the redundant edges from occurring. Like Graph-GA, the graph generated by NND is usually not fully connected. As Fig. 6A shows, NND generates seven sub-graphs, each representing one cluster, which is perfectly consistent with the underlying clustering structure. Besides, each sub-graph (also an IT) has one root node (in red circle).

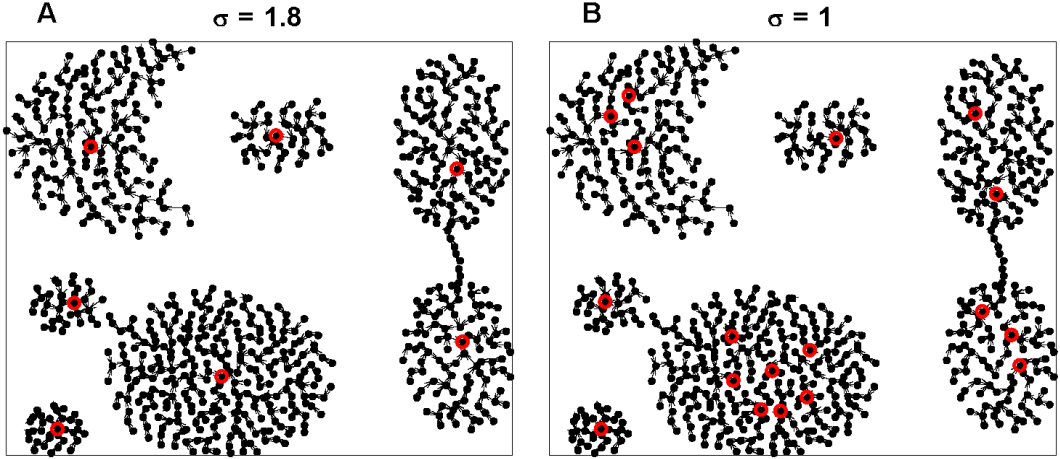

**Fig. 6. The graphs constructed by NND with different values of parameter σ.** Left: well partitioned result; right: over-partitioned result. The points inside the red circles are the root nodes (or cluster centers) of each separate sub-graph (also an IT).

Although, compared with ND, NND can complete the task once and for all, without the additional requirement of removing the redundant edges, NND doesn't become more powerful than ND. NND is not always as perfect as what it looks in Fig. 6A. Because, it is easy to find that NND is quite alike DENCLUE, Mean-shift and Graph-GA, and thus NND is also severely affected by the "ripple noise", as revealed by the "over-partitioning" result in Fig. 6B. Nonetheless, by the above perspective, we found another value of NND, that is,

> ***NND** can serve as the **tie** between the new methods (e.g., DG, ND, the following H-NND, and the proposed method, D-NND, in this paper)*

*and the previous ones (e.g., DENCLUE, Mean-shift and Graph-GA).*

### 1.7 Hierarchical Nearest Neighbor Descent (H-NND)

Subsequently, we proposed another method, called the Hierarchical Nearest Neighbor Descent (H-NND) (10), in which the "descending" process was divided into 2 stages. The first stage was executed by NND only for the non-extreme nodes $\Psi - Y$ (see the definition for $Y$ in the last subsection, i.e., the root nodes in red circles in Fig. 6); the 2nd stage was executed by ND for the extreme nodes $Y$. The expressions can be written as,

$$\text{The 1st stage (NND):} \quad I_i = \arg\min_{j \in \eta_\theta^i} d_{ji} \cdot \varphi(-P_{ji}), \text{ for } i \in \Psi - Y \quad (12)$$

$$\text{The 2nd stage (ND):} \quad I_i = \arg\min_{j \in Z} d_{ji} \cdot \varphi(-P_{ji}), \text{ for } i \in Y. \quad (13)$$

Like ND, H-NND also generates the IT structure (Fig. 7) while the redundant edges could become more salient and thus could make the repair process (i.e., removing the redundant edges) become much easier and more reliable. In both cases (A and B) in Fig. 7, by removing the six longest edges, the well clustered results would be obtained without the "over-partitioning" problem.

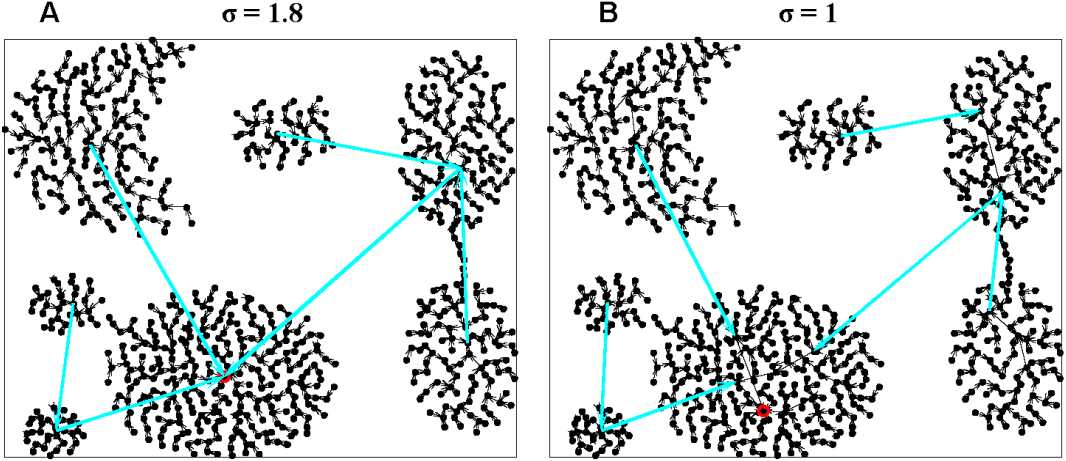

**Fig. 7. The IT constructed by NND.** In each case (left or right), the redundant edges (cyan) are obviously the six longest ones.

## 2 Motivation and Idea of this paper

Despite that DG, ND and H-NND could largely reduce the risk of the under-smoothed density or potential to the ultimate clustering results, while the risk of the over-smoothed case to the clustering result still exists. In other words, these methods could still be affected by the kernel bandwidth, despite the "safe range" for the bandwidth has become relatively larger. Theoretically, one can adjust the bandwidth, but judging whether the current bandwidth leads to a better clustering result is nontrivial. Besides, there are cases, e.g., *multiple scales (or resolution) in different clusters, or unbalanced element numbers in different clusters*, that one fixed kernel bandwidth cannot well reveal the underlying density. That is why we previously proposed a visualization method, called IT-map (27), to help set a more reliable bandwidth or different bandwidths for different clusters by using the divide-and-conquer strategy under the supervision of users. Besides, one can expect a kernel density estimation method that could make the bandwidth adaptive to the local distributions of data points.

Nevertheless, is it possible that the bandwidth problem (both the associated over-smoothed and under-smoothed risks for the density estimation) could be solved all by the density-based clustering method itself rather than the help of IT-map or adaptive kernel density estimation methods? In order to fulfill this goal, we should first answer such question: what is the cause for the bandwidth problem? In our perspective, the answer lies in the fact that the two processes involved, i.e., *density estimation* and *clustering*, are separate and performed in a serial way (as the questions in Section 1.1 reveals) for those density-based clustering methods DG, ND and H-NND (the latter two can also be viewed as density-based methods), and consequently the clustering process largely relies on the performance of the density estimation.

Therefore, in this paper, we will propose a density-based hierarchical clustering method, called the Deep Nearest Neighbor Descent (D-NND), in which the above two processes will **interplay.** This is fulfilled by making full use of the hierarchical strategy and the merit of NND. Specifically, in each layer of the hierarchy, the density (actually we still use the *potential* form in D-NND) on certain nodes are updated based on the discovered cluster structure, and in turn, the updated density estimation will be used to renew the cluster structure. The density in each layer is estimated based on the local information, whereas as the layer number increases, the magnitudes of the estimated potentials on the sample nodes in the higher layers could gradually grasp the global density distribution in the dataset. By this way, the proposed method could be adaptive to the multi-resolutions of the different clusters.

*D-NND* is expected to not only largely reduce the negative effect made by the under-smoothed potential estimation, a property inherited from ND and H-NND, but also largely reduce the risk of over-smoothed case. In effect, the proposed method could appear insensitive to the parameters in considerably large ranges of values.

## 3  Method

### 3.1  The details of the proposed method: D-NND

D-NND contains two stages: the bottom-up and top-down stages.
**Bottom-up stage** (making all data points organized into the IT):

As summarized in Table 1, D-NND takes as input the distance $d_{i,j}$ between any pair of data points $i, j \in \Psi = \{1,\cdots,N\}$. At the beginning, all points are of zero potential (Note that in order to use NND, we use the term potential $P_i$ rather than density $\rho_i$, whereas $P_i = -\rho_i$), and $X = Y = \Psi$, where $X$ and $Y$ respectively denote the input and output dataset in each layer.

Each layer contains 5 steps. First, the neighbor nodes (denoted as $\eta^i$) of each node $i \in X$ are identified by constructing the Neighborhood Graph such as $k$-Nearest-Neighbor graph[7] in which each node selects the $k$ nearest nodes as its neighbors. Then, the local potential $P_i$ on each node $i \in X$ is computed via summing the dissimilarities between point $i$ and all its neighbors plus the history potential of node $i$ in the last layer. Then comes NND method, which is divided into two steps 3 and 4. In Step 3, the *candidate* parent node set $J_i$ of each node $i \in X$ is defined, and the nodes (corresponding to the locally extreme points) with null $J_i$ is denoted as dataset $Y$. In steps 4, the parent node $I_i$ of each node $i \in X - Y$ is identified. The root nodes in the dataset $Y$ will serve as the input for the next layer. The last layer occurs in the time when there is only one root node (denoted as node *r*) in $Y$.

*In conclusion*, given the input nodes in *X*, each layer functions to: (i) update the potentials of the nodes in $X$; (ii) identify the root nodes $Y$; (iii) identify the parent nodes for the remaining nodes in $X - Y$.

As illustrated in Fig. 8A, if we connect each node to its parent node and view the points in $Y$ as the root nodes (the red points in Fig. 8A), in effect, data points are nested layer by layer (Fig. 8A, from left to right) and the number of root nodes is reduced layer by layer, until all the data points are organized into the fully connected IT data structure with only one root node left (Fig. 8A layer 3). For the IT data structure, the data points in the original input $\Psi = \{1, \cdots, N\}$ correspond to the nodes in it. Each node *i* (except node *r*) and its parent node $I_i$ respectively defines the start and end nodes of one directed edge ($i$, $I_i$) and the distance $d_{i,I_i}$ between them defines the edge length $W_i$. Note that, each directed edge ($i$, $I_i$) in IT is the only directed edge, denoted as $e_i$, started from node *i*. In other words, the start node of each edge in IT can serve as the unique identifier of each edge. This is the basis for the Decision-Graph-Cut in the Top-down stage.

---

[7] Also, some non-parametric neighborhood graph can be used, such as the Delaunay Graph.

## Table 1. The bottom-up stage of D-NND

**Input:** distance $d_{i,j}$. $i, j \in \Psi = \{1, \cdots, N\}$.

**Procedure:**

0. Initialize: $P_i = 0, i \in \Psi$; $X = Y = \Psi$.

1. Identify the neighbor nodes $\eta^i$ of each node $i \in X$.

2. Estimate the potential on each node $i \in X$:

$$P_i = P_i + \sum_{j \in \eta^i} D(d_{i,j}); \quad // D(x) = x \text{ or } -e^{-x/\sigma}.$$

3. Identify the candidate parent nodes of each node $i \in X$:

$$J_i = \{j \mid P_j < P_i, j \in \eta^i\}, \text{ and denote } Y = \{i \mid J_i = \varnothing\}.$$

4. Determine the parent node of each node $i \in X - Y$:

$$I_i = \arg\min_{j \in J_i} d_{i,j}, \text{ and denote } W_i = d_{i,I_i}.$$

5. If $|Y| \neq 1$, then $X = Y$, repeat steps 1~4; otherwise, $I_i = i, W_i = -\inf$.

**Output:** $I_i$ and $W_i$. $i \in \{1, 2, \cdots, N\}$.

\* $|Y|$ denotes the number of data points in $Y$; IT is featured by $I_i$ and $W_i$.

**Top-down stage** (removing the redundant edges):

In order to divide the dataset into groups (i.e., clustering), the fully connected IT requires to be divided into pieces (each representing a cluster) by removing the redundant edges in IT.

Pleasingly, like the IT structures constructed by ND (Fig. 5C) and H-NND (Fig. 7), the IT generated here (Fig. 8A, layer 3) also shares the good property that the redundant edges (cyan) are distinguishable from other edges, and thus they are not hard to be determined, as Fig. 8B shows, the redundant edges can be easily determined by two different methods: ***E-cut*** and ***Decision-Graph-Cut***. *E-cut* is the plot of the lengths of all edges in IT in decreasing order (Fig. 8B, left). *Decision-Graph-Cut* (see also section 1.5) features each directed edge $e_i$ in IT by two associated variables, its edge length and the potential of its identifier (i.e., its start node), as the right image in Fig. 8B shows, the pop-out points (interactively determined in the red box) could rightly correspond to the redundant edges in IT. As Fig. 8C shows, the above two methods lead to the same and almost perfect clustering result.

In fact, besides *E-cut* and *Decision-Graph-Cut*, we have previously devised other effective methods (with different advantages) to remove the redundant edges in IT, e.g., **IT-map** (27), **IT-Dendrogram** (28), **G-AP** (29) and **SS-cut** (8).

For simplicity, we will still use the E-Cut and Decision-Graph-Cut only in the following experiments to demonstrate the power (effectiveness and insensitivity to the parameters) of the proposed method D-NND. The saliency of the redundant edges and the diversity of the edge removing methods could guarantee the good performances of clustering results.

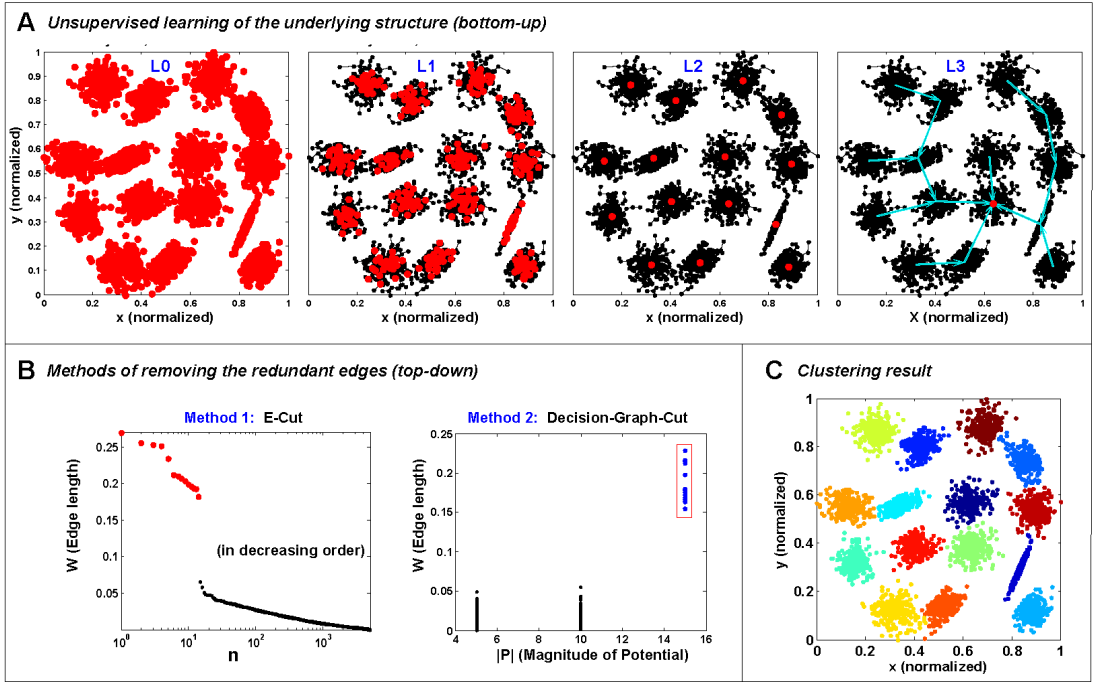

**Fig. 8** An illustration for D-NND on a dataset with *N* = 5000 data points. **(A)** From left to right, as the layer increases from 0 to 3, the number of the root nodes (red) was successively reduced from 5000, 418, 15 to 1, and at the same time, the whole data points were gradually organized into the fully connected graph, i.e., the IT (the rightmost image). For clustering purpose, the redundant edges (cyan) in IT require to be removed, which are, however, obviously distinguishable (*with longer lengths at least*) from the other edges and thus can be easily determined. **(B)** Different methods to determine the redundant edges in IT. Left (*E-Cut*): the plot of the lengths (in decreasing order) of all edges. The *K* longest edges that need to be removed can be easily determined according to the plot by setting a threshold between points with a large gap or just by counting the number of the points with saliently large values (in red). Right (*Decision-Graph-Cut*): the interactively determined pop-out points (blue) correspond to the start nodes of the redundant edges. These two methods lead to the same clustering result **(C)**. The almost perfect clustering results demonstrate the effectiveness of the whole process. In this illustration, we set $k = 5$, $\sigma = 100$.

## 3.2 Parameters

The proposed method has at most 2 parameters, one is the neighbor number $k$, if the $k$-Nearest-Neighbor graph ($k$-NN) is used to define the neighborhood relationship in step 1; the other is a kernel-like parameter σ, if the exponential function is used to compute the dissimilarity in step 2. We will show in the experiments that, the proposed method is not sensitive to both $k$ and σ in an extremely large range. In fact, the parameter could now function like the "fine-tune knot", which is, however, a good thing from the technical point of view.

## 4  Experiments

We first tested three 2D datasets (Fig. 9) from (30), (7), (31), respectively, using large range of values for both $k$ and σ. For the 1st dataset: $k$ was varied from 5 to 50 while σ = 1 or 10000. For the 2nd dataset: $k$ was also varied from 5 or 50 while σ = 1 to 10000. For the 3rd dataset: $k$ was varied from 5 to 40 while σ = 1 to 10000; Note that, for the first dataset, the data in each dimension were normalized to [0 1] like what we did in Fig. 8. In all cases, there are points popping out in the Decision Graphs and the corresponding clustering results are all consistent with our visual perception as Fig. 9 shows. In fact, it can be seen that the same clustering performances can also be achieved by using the E-Cut, since the edge length variable alone (vertical axis in Decision Graphs in Fig. 9) is enough to distinguish those pop-out points (or the corresponding redundant edges). In order to further demonstrate this, we did such experiments on the first dataset that $k$ varies from 2, 10 to 40 and σ varies from 0.1, 100 to 10000. In each case, the 14 longest edges in the IT structures were removed, which results in 15 clusters. By comparing the clustering assignments of all results with the benchmark data, the average error rate for the clustering assignments is almost negligible: 0.0057 ± 0.0006 (mean ± standard deviation). In other words, all of the clustering results are almost perfect when both $k$ and σ vary in large ranges of values.

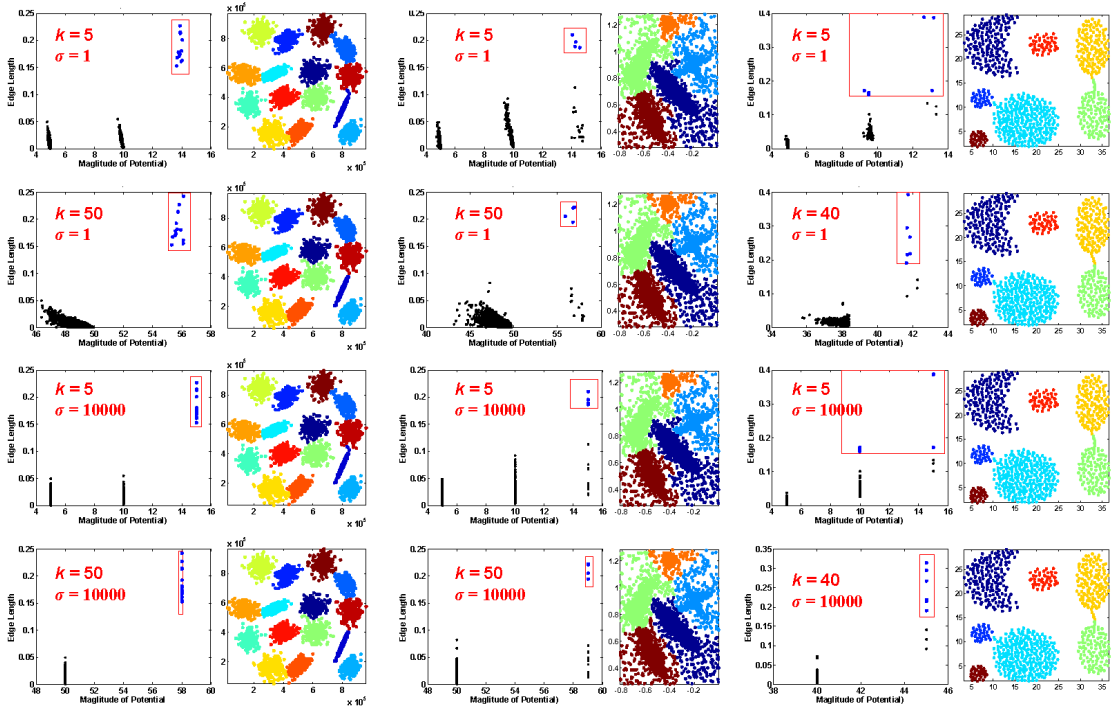

**Fig. 9. Tests on three 2D datasets.** The parameters $k$ and σ, especially σ, vary in large ranges of values for each dataset. Euclidean distance was used to compute the pair-wise distance.

Then, we tested a set of (*five*) high-dimensional datasets from (32), each containing $N = 1024$ data points (sampled from $M = 16$ Multivariate Gaussian functions), whereas the dimension $d$ varies from 32, 64, 256, 512, to 1024. For each dataset, we chose two largely different values for $k$ (= 5 or 500), together with two extremely different values for $\sigma$ (= 1 or 100000). We used E-Cut for all cases. The plots of the edge lengths are shown in Fig.10. In each plot, there is a salient gap between the determined number of largest edge lengths (*in red*) with the rest ones by setting an appropriate threshold in the gap. Almost all of the corresponding clustering results are perfect. The clustering error rate for all cases are 0 and the cluster numbers for most of them are consistent with the underlying number (M = 16) of groups, except few of them with slightly larger cluster numbers, being either 17 or 18. However, we found that the extra cluster(s) contains only one point, actually being regarded as the outlier.

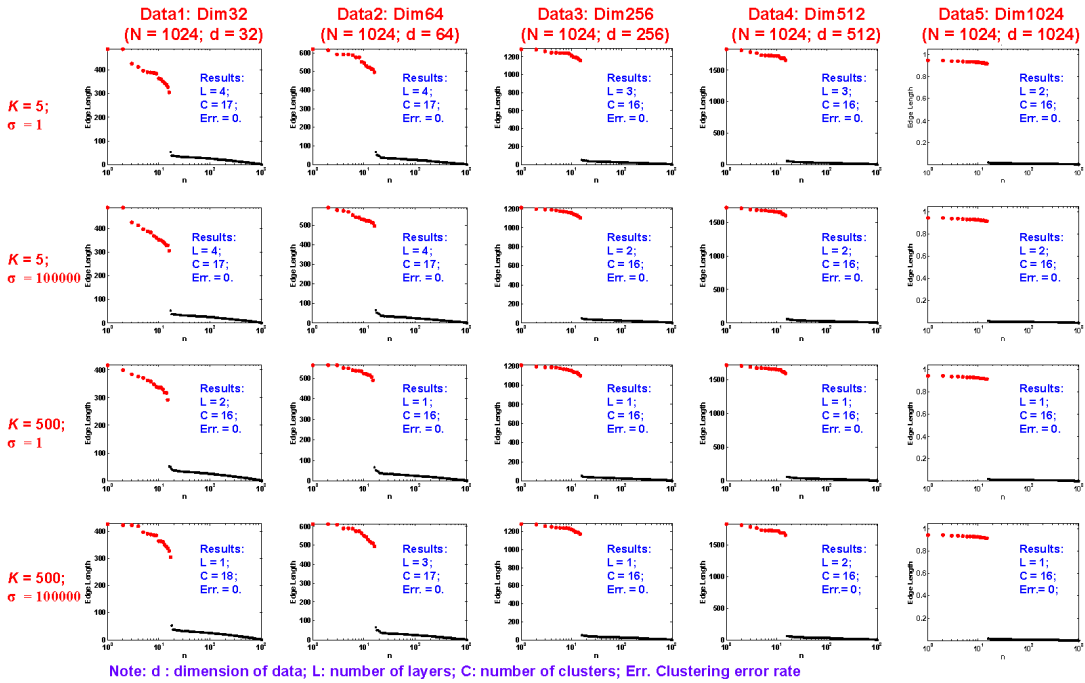

**Fig. 10. Tests on five high-dimensional datasets** (from left to right, $d$ = 32, 64, 256, 512, 1024). The parameters $k$ and $\sigma$ vary in large ranges of values for each dataset. Euclidean distance was used.

We also tested the United State Poster Service (USPS) digit number dataset, which contains $N = 11000$ grayscale handwrite digits. Each digit is a $16 \times 16$ image, treated as a 256-dimensional vector in the test. Here, we used the Decision-graph-Cut, expecting to reach small error assignment with small cluster number. We also selected five significantly different values for $k$ (= 2, 5, 10, 20, or 50) and two extremely different values for $\sigma$ (= 1 or 100000). The testing results for all cases are shown in Fig. 11.

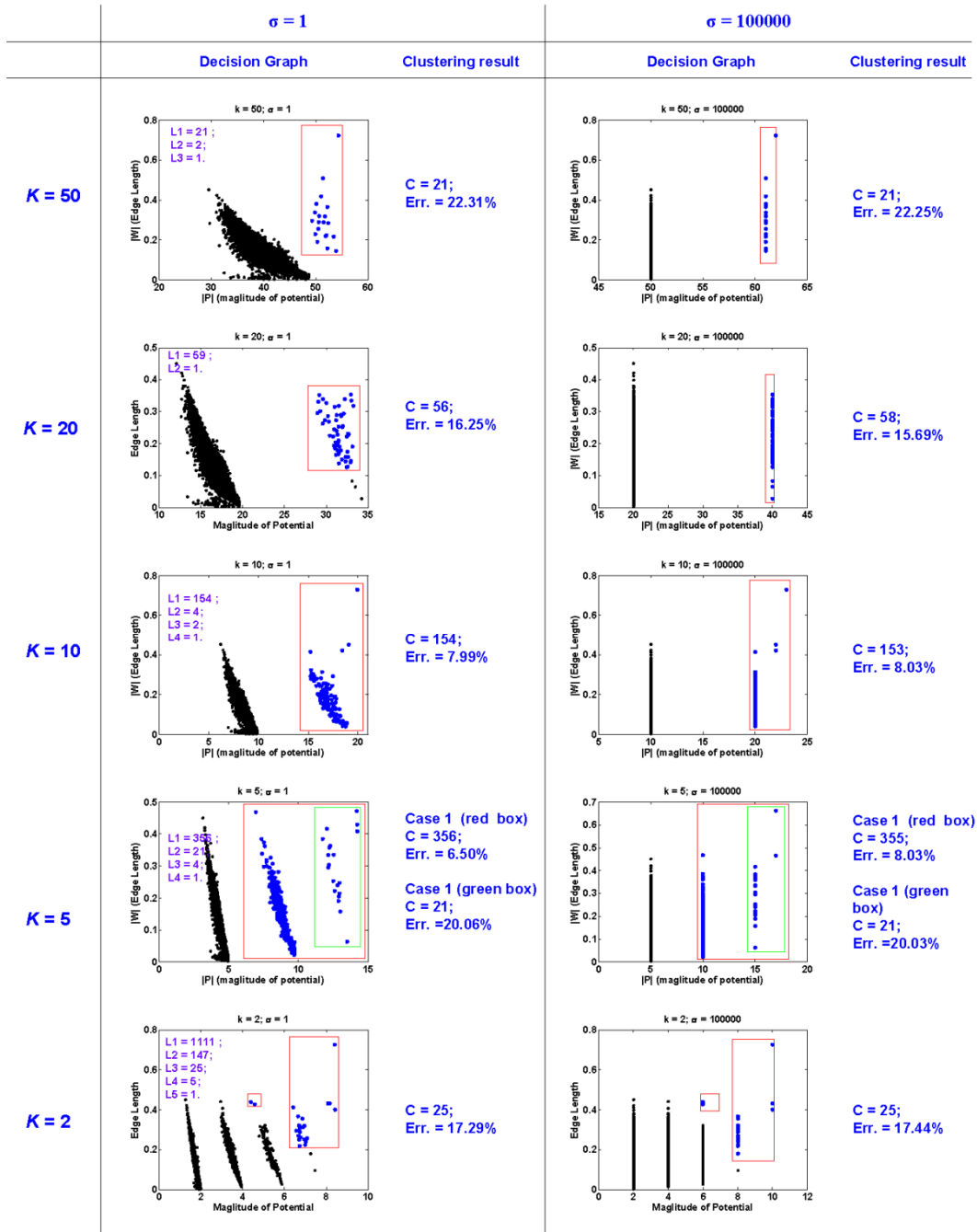

**Fig. 11. Tests on the USPS digit number dataset with large ranges of values for parameters *k* and σ, especially σ.** Cosine distance was used.

## 5   Conclusions

In Section 1, we first summarized two general questions (i.e., the questions Q1 and Q2) of the density-based clustering and accordingly introduced seven density-based clustering methods, i.e., DENCLUE, Mean-Shift, Graph-GA, Rodriguez and Laio's DG, and the methods proposed by us before (i.e., ND, NND and H-NND), and at the same time we introduced (i) four problems (P1~P4) of DENCLUE or GA; (ii) how other methods partly solve those problems; (iii) the relationship between DG and ND: DG can be viewed as the Nearest Ascent (NA), the counterpart of ND; (iv) the relationship between the newly proposed novel methods (DG, ND and NND, since 2014) and previous methods (DENCLUE, Mean-Shift and Graph-GA): NND could serve as the tie for them. In Section 2, we explained the motivation (a density-based clustering method that can solve the density estimation problem all by itself) and the main idea of this paper (taking the two processes, density estimation and clustering, interplay). In Section 3, we introduced the proposed method D-NND in details. In Section 4, we showed in the experiments that D-NND has not only the strong capability of discovering the underlying cluster structure but also the remarkable reliability due to its insensitivity to parameters in large ranges of values.

Although NND can only avoid the first three problems of GA, via the proposed hierarchical learning framework (unsupervised), the last problem of GA is also largely solved by the proposed method D-NND. This endows D-NND with such advantages: non-iterative, unconstrained by the attributes of the data, and insensitive to kernel bandwidth, or in other words, the efficiency, general meaning, and reliability. Besides, D-NND also shares the *seven* general merits (in the beginning of Section 1) of most of the density-based clustering algorithms.

# 6  Discussions

### 6.1  Bottom-up exemplar election.

Like the affinity propagation (AP) (33), the bottom-up stage of the proposed D-NND can also be interpreted as the process of selecting exemplars (or representatives). In this prospective, Fig. 8A can be analyzed like this: at the beginning, all points are viewed as the exemplars of themselves. After comparing among the exemplars, several of them (local density points) will continue to be the exemplars of the other exemplars in the next layer. This process proceeds in the higher layers until one exemplar becomes the exemplar of all the remaining exemplars in certain layer. Thus, the whole bottom-up stage can be roughly viewed as a hierarchical exemplar election system. Since in each layer only the local information is considered, this could make the selection in each layer to be local optimum. And as the layer increases, the exemplars could actually represent large ranges of areas of points from a global perspective. In other words, the selection will gradually become global optimum in the higher layers.

### 6.2  "Center-biased" trend for the exemplars guarantees the saliency of the generated redundant edges.

Before being replaced (in the higher layers) by the exemplars in other clusters, the exemplars in the same clusters will gradually evolve to the centers of the clusters, forming a "center-biased" phenomenon (Fig. 8A), which guarantees the saliency of the generated redundant edges (Fig. 8A, layer 3), since each redundant edge will approximately start from the center of one cluster and end in the center of another cluster. This is in stark contrast to the case for the MST in which the redundant edges usually occur in the marginal areas of clusters and thus usually being hard to be determined. Note that, Fig. 8A shows the case that all undesired edges are generated simultaneously in the last layer, but there is also case that the redundant edges (cyan) are generated in different layers as Fig. 12 shows.

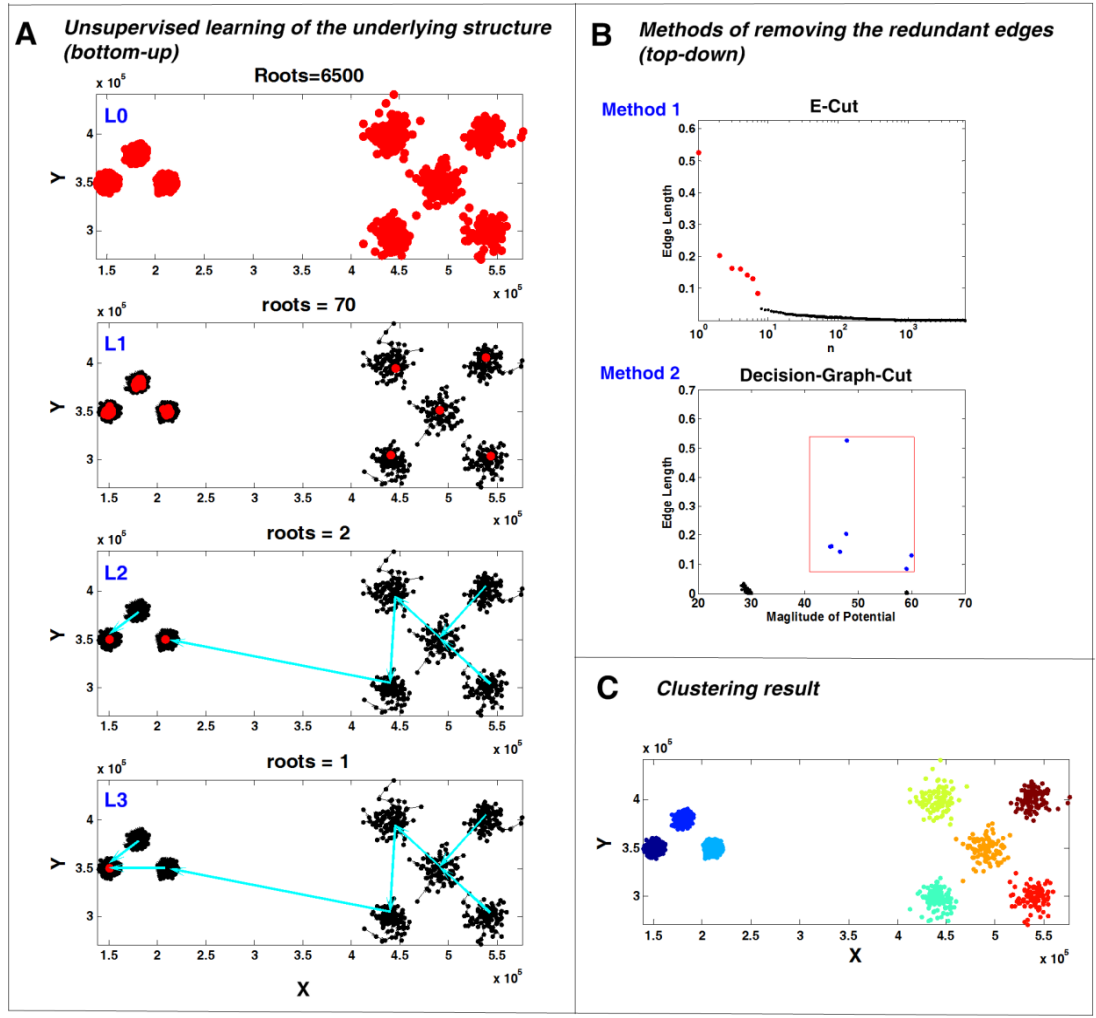

**Fig. 12.** Test on the unbalanced dataset[8]. $k = 30$, $\sigma = 1$;

### 6.3 Insensitive to parameters.

On the one hand, since density estimation in each layer is based on the neighborhood relationship, the kernel is constrained in a local sphere and thus, no matter how large the kernel-like parameter σ is, it makes very litter effect on the global distribution and thus the over-smoothing density phenomenon is largely avoided, provided that the neighborhood relationship is to some degree guaranteed (for the nonparametric neighborhood graph such as Delaunay Graph, this is always valid; and for $k$-NN graph,

---
[8] Downloaded from http://cs.joensuu.fi/sipu/datasets/

*k* should not be too large). On the other hand, such local constraint for the density estimation would make the under-smoothed density problem (in the lower layer) even worse. For DENCLUE (or GA), Mean-Shift, Graph-GA, and NND, this would mean a severer over-partitioning result. However, as revealed by the experiments, this under-smoothed case is not so troublesome to D-NND either, due to the hierarchical procedure (which gradually improves this situation) and the strategy of redundancy design (which largely reduces the effect of the ripple noise). Overall, the prevention for the over-smoothed density estimation and the ability to tackle with the problem (ripple noise) brought by the under-smoothed case make the proposed method insensitive to the parameters (especially the kernel-parameter) in large ranges.

### 6.4 NND vs. GD: particularity vs. generality

For one thing, hierarchical strategy largely solves the over-partitioning problem of NND. For another, the role of NND is fully played and revealed, as a significant element in each layer. The reasons are analyzed as follows.

Suppose that we use in each layer the Gradient Descent (GD) (i.e., the counterpart of GA), rather than NND, then the first two problems (i.e., Q1 and Q2 in Section 1) of GA will also exist here. And as the layer number increases, the overall problems will be more nontrivial in this hierarchical system. In contrast, given the estimated potentials on nodes, NND brings in no additional problem, no matter how many layers the hierarchy contains.

***So, what makes it possible that NND could be an alternative of GD in this task?*** In our perspective, GD provides a general solution to the problem of locating the maximum points in the density function, i.e., the traditional optimization problem. In contrast, NND grasps the "***particularity***". This "particularity" mainly refers to the fact that the value set of the independent variable for the assumed underlying density function in this particular problem here is ***finite***, ***discrete*** and ***known*** (i.e., the input dataset). For this reason, a graph (or map) could be constructed after the first independent hopping of each point to its parent node. In fact, this "particularity" can also explain why previously there could also exist other alternatives of GA, i.e., Mean-Shift and Graph-GA.

In fact, the "***steepest pursuit***" behind GD (***or GA***) is not necessary in terms of computation time in this task. The reason is that although GD (or GA) lets each point choose the steepest path to reach the density peak, each step of approaching the density peak involves the equivalent computation. In comparison, for NND, only the first step involves a certain degree of computation, but the computation time by the remaining steps are almost negligible due to the pre-specified paths on the constructed graph. Even compared with Graph-GA, the graph-based approximation of GA, the

"steepest pursuit" also makes no considerable advantage in computation time. Admittedly, the path to the root node for each node in the graph constructed by Graph-GA is generally nearer to the "steepest path" and thus contains less directed edges, as compared with NND. This is illustrated in Fig. 13. For the same point (in blue), the path in the left image (Graph-GA) obviously contains less edges than that in the right image (NND). And overall, the number of the total edges generated by Graph-GA is less than that by NND. Nevertheless, this brings Graph-GA negligible advantage in terms of the computation time in this graph-based searching, as compared with the similar case in NND, since, for NND, the computation time for such *kind* of graph-based search could already be negligible in general, revealed in Table S2 in (8), where[9], for instance, the computation time on a dataset with $N$ = 8124 data points costs no more than 0.005s. Note that although the graph in (8) is not constructed by NND, the way of searching the root nodes is of the same process as here.

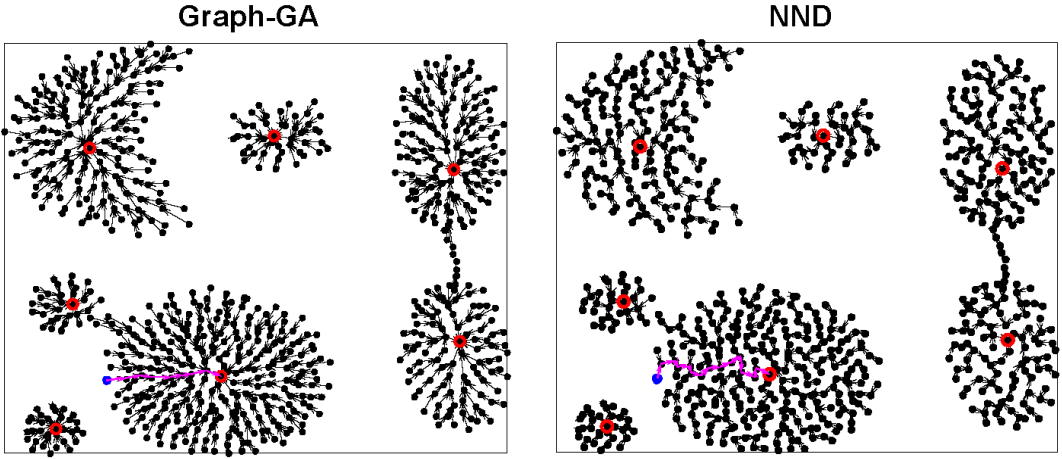

**Fig. 13. Comparison between Graph-GA and NND in terms of the path.** The graph generated by Graph-GA contains less edges. $\sigma$ = 1.8, $k$ = 10.

Beside, this "steepest pursuit" is also not more efficient in terms of the parameter in this task. The reason can be revealed by comparing NND and Graph-GA. Compared with Graph-GA, NND is less sensitive to the parameter $\theta$ due to the constraint of the "nearest" requirement. This has been partly shown in H-NND for its insensitivity to the neighborhood parameter in a large range, and this is also fully revealed by the stark comparison between NND (left column) and Graph-GA (right column) in Fig. 14 with different values of the neighborhood parameter $k$. Note that, we use the $k$-NN graph for both of them to define the neighborhood relationship. Besides, in the literature of Graph-GA (4), the authors first estimate the density based on the neighborhood relationship. Here in Fig. 14, in order to compare NND and Graph-GA, we assume that for both NND and Graph-GA, the underlying densities have already been well estimated as what Fig. 1B shows. We can see that, compared with the corresponding result in Fig. 13 ( $k$ = 10 ), when $k$ is increased to 20 in Fig. 14, there is no significant change for NND, whereas the problem (denoted by the red circle) starts to occur for Graph-GA. When $k = 200$ or 787, only four or seven undesired edges (in cyan) are falsely generated by NND, whereas the corresponding results for Graph-GA are severer. Besides, for NND, by removing the largest four or seven edges, good clustering results will be obtained, whereas, for Graph-GA, when all the redundant edges are successfully removed, this would lead to severe over-partitioning clustering results. Besides, note that when $k = N − 1$ (= 787 here), NND becomes ND, which is obviously not a too bad circumstance.

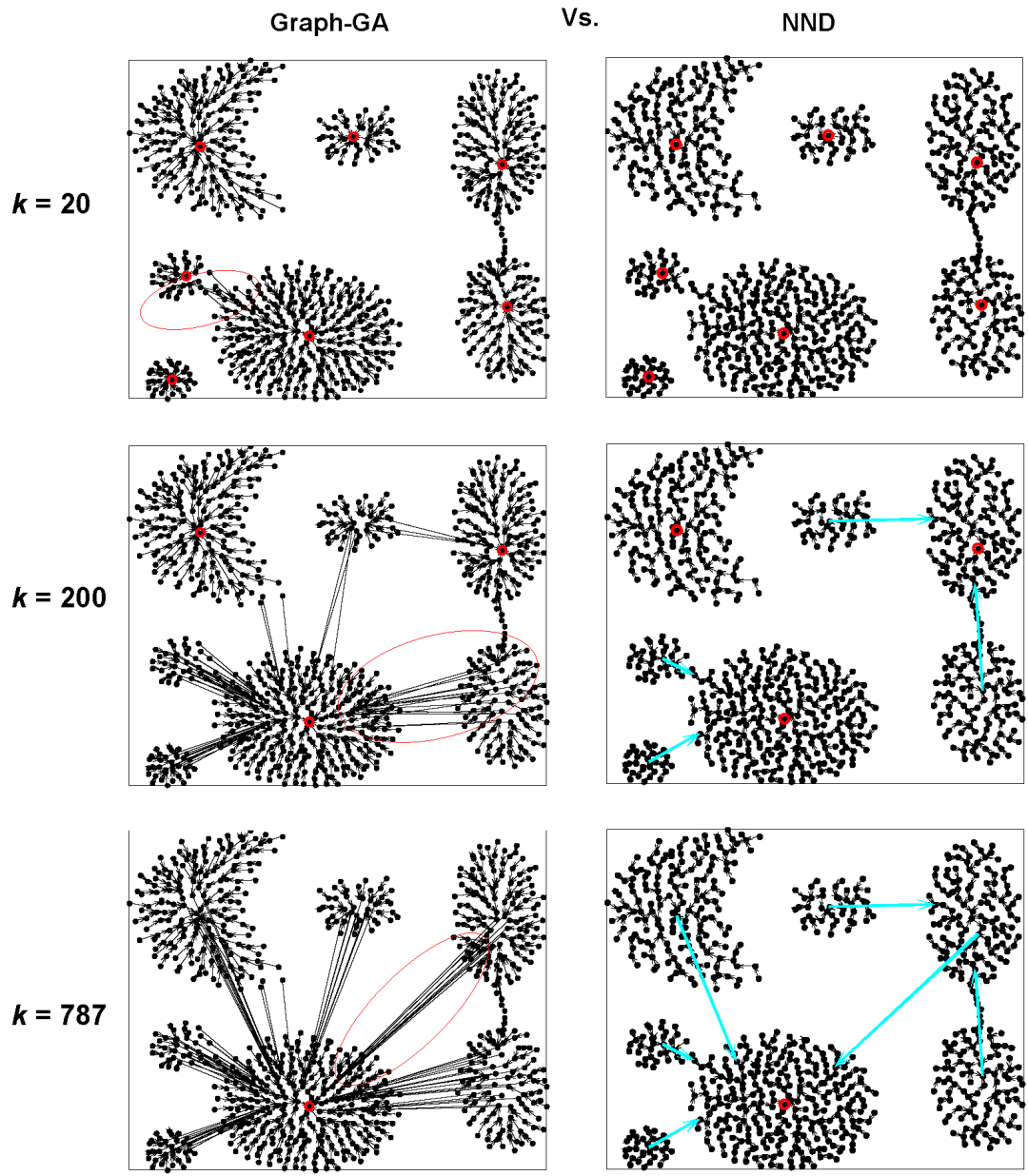

**Fig. 14. Comparison between Graph-GA and NND in terms of parameter.** NND presents much better performance while the value of $k$ is increased. $\sigma = 1.8$

### 6.5 The Fault-tolerant design.

Unlike DENCLUE, Mean-Shift, Graph-GA and NND, D-NND does not seek to directly get the clustering result once and for all. Instead, it follows such ***Fault-tolerant design (inherited from our previous methods ND and H-NND)***:

> *how to have data effectively organized first* (*referring to constructing the IT structure here*) *with the tolerance of making "mistakes"* (*referring to the redundant edges in IT*), and *then seek to "repair them"* (*referring to removing the redundant edges here*).

In our perspective, although hierarchical strategy makes H-NND generate more distinguishable redundant edges and D-NND more robust to the parameters compared with ND, the most contributing factor for D-HHN, together with ND (or Rodriguez and Laio's DG) and H-NND, is this Fault-tolerant design. The reasons are analyzed as follows.

> *Unlike DENCLUE, Mean-Shift, Graph-GA and NND, all the particular nodes, i.e., the modes or the density peaks (either valid or invalid), in ND, H-NND and D-NND do not have the "privilege", and consequently the redundant edges for the "valid modes" would be produced. However, the tolerance to such kind of "mistakes" (referring to the redundant edges) also endows them with the resistance to another kind of "mistakes" (referring to the invalid modes in the "ripples noise"), and thus the risk of generating the over-partitioned clustering results is largely reduced for them. Also due to the guideline of the fault-tolerant design, the proposed D-NND can further reduce another risk, i.e., the risk of generating the under-partitioned clustering results, by using another novel implementation of constructing IT.*

Besides, due to the Fault-tolerant design here, the IT-based clustering family is enriched by a variety of methods (e.g., different methods to construct IT: ND, H-NND, D-NND; and different methods to remove the redundant edges in IT: E-Cut, Decision-Graph-Cut, IT-map, G-AP, etc.). Similarly, in our opinion, the Fault-tolerant design could also be used to explain the enrichment of the traditional link-based hierarchical clustering (***L-HC***) methods in which different methods (e.g., single linkage, complete linkage, average linkage) have been proposed to construct the Dendrogram.

Despite the similarity in terms of the Fault-tolerant design, however, compared with L-HC, the advantages of the IT-based clustering are twofold: (i) for this Fault-tolerant design, one thing vital is that how the "mistakes" are designed or will be generated, since the more distinguishable the mistakes are, the easier and more reliable the following repair methods could be. For ND, H-NND and the proposed me-thod D-NND, this is just their biggest advantage, since the redundant edges could in general be distinguishable, in sharp contrast to the traditional L-HC (especially the single-linked one, which is generally equivalent to MST). (ii) Besides the saliency of the redundant edges, the diversity of the repair methods is another advantage for the IT-based clustering methods compared with L-HC, since a set of effective repair methods (as mentioned in the above paragraph) can be used to help remove the error links in this particular graph structure, IT, also in sharp contrast to the only choice (i.e., the Dendrogram) for L-HC.

It seems that this Fault-tolerant design could be a quite efficient way to reach a certain degree of robustness while at the lowest cost and using a simpler system.

**6.6 Think globally, while learning locally in hierarchy.**

Like the idea ("*think globally, fit locally*") in the famous dimensionality reduction method locally linear embedding (*LLE*), here we also use the local information to estimate the density (characterized by *potential* in this paper), despite it is a global feature. Nevertheless, in this special problem, we also combine such local estimation with the hierarchical strategy so that, as the layer increases, the magnitudes of the estimated potentials on the points in higher layer can gradually reveal more and more global density features. Also, since, in each layer, the potential is updated based on the local neighborhood relationship (which is relatively reliable) instead of the single-scaled kernel bandwidth (in fact the effect of the bandwidth has been largely constrained, see Section 6.3), the density estimation could be in effect adaptive to the multiple scales in the clusters.

In fact, in our opinion, the successes of such methods as Segmentation by Weighted Aggregation (*SWA*) (34) and Tree Preserving Embedding (*TPE*) (35) in solving the similar multi-resolution challenges in other unsupervised tasks (image segmentation and dimensionality reduction, respectively) are also largely benefited from the similar idea reflected in this D-NND, that is, *Thinking globally, while learning locally in hierarchy*.